\documentclass[conference]{IEEEtran}
\IEEEoverridecommandlockouts
\usepackage{cite}
\usepackage{amsmath,amssymb,amsfonts}
\usepackage{mathtools}
\usepackage{algorithmic}
\usepackage{graphicx}
\usepackage{todonotes}
\graphicspath{ {./images/} }
\usepackage{textcomp}
\usepackage{xcolor}
\def\BibTeX{{\rm B\kern-.05em{\sc i\kern-.025em b}\kern-.08em
    T\kern-.1667em\lower.7ex\hbox{E}\kern-.125emX}}
\usepackage{booktabs}
\usepackage{nidanfloat}
\usepackage{hyperref}
\usepackage{amssymb}
\usepackage[utf8]{inputenc}

\usepackage{svg}
\usepackage{tabularx,ragged2e} 
\newcolumntype{C}{>{\Centering\arraybackslash}X}

\begin{document}

\title{Deep Variational Autoencoder with Shallow Parallel Path for Top-N Recommendation (VASP)}

\author{\IEEEauthorblockN{Vojt\v{e}ch Van\v{c}ura}
\IEEEauthorblockA{\textit{Faculty of Information Technology} \\
\textit{Czech Technical University in Prague}\\
Prague, Czech Republic \\
vancurv@fit.cvut.cz} 
\and
\IEEEauthorblockN{Pavel Kordík}
\IEEEauthorblockA{\textit{Faculty of Information Technology} \\
\textit{Czech Technical University in Prague}\\
Prague, Czech Republic \\
pavel.kordik@fit.cvut.cz \\
0000-0003-1433-0089}
}

\maketitle

\begin{abstract}
Recently introduced EASE algorithm presents a simple and elegant way, how to solve the top-N recommendation task. In this paper, we introduce Neural EASE to further improve the performance of this algorithm by incorporating techniques for training modern neural networks. Also, there is a growing interest in the recsys community to utilize variational autoencoders (VAE) for this task. We introduce deep autoencoder FLVAE benefiting from multiple non-linear layers without an information bottleneck while not overfitting towards the identity. We show how to learn FLVAE in parallel with Neural EASE and achieve the state of the art performance on the MovieLens 20M dataset and competitive results on the Netflix Prize dataset.
\end{abstract}


\section{Introduction}
  
    With the increasing amount of information on the Web, Recommender Systems (RS) are an important way to overcome infobesity.  On the other hand, companies like NetFlix, Youtube, Amazon, or Google are making significant revenues from recommendations\footnote{According to \cite{Zhang2019a} 80\% movies on NetFlix, 60\% videos on Youtube are watched based on recommendations; recommendations \ are responsible of 35\% sale revenues on Amazon \cite{Symeonidis2016}}. Thus, RS are gaining more attention over the past two decades.
    
    As online companies grow, RS have to scale to millions of active users and millions of items. Speed of training and recall are also increasingly important as available content often change dynamically and RS need to be able to react in real-time. 
    
    Proper evaluation of RS is also increasingly important topic as offline evaluation is often biased predictor of the online performance \cite{rehorek2018comparing}. For offline evaluation, the recsys community shifted towards Top-N approaches \cite{karypis2001evaluation} as evaluating the performance based on root mean squared error on top of a predicted rating matrix can be very misleading.    
    
    In the Top-N recommendation scenario\cite{Cremonesi2010}, RS is recommending $N$ most relevant items for every user. This is the typical case in various domains, including media, news, or e-commerce.
    
    Various approaches have been proposed for solving the Top-N recommendation task, including collaborative filtering with matrix factorization to give an example. Recently, sparse-data autoencoders \cite{vincent2008extracting,ng2011sparse,Sedhain2015,Liang2018,Kim2019,Shenbin2020} gained much attention and were providing state-of-the-art results in solving this task.
    We examined various proposed models, including denoising autoencoders, variational autoencoders, and shallow autoencoder called the EASE \cite{Steck2019}, which despite being a simple linear model, is providing competitive and explainable results while addressing the biggest problem with sparse autoencoders: overfitting towards identity.
    
    Inspired by \cite{Cheng2016}, our motivation was to build a RS model, that is as elegant and explainable as EASE while leveraging the potential of deep autoencoders to model complex nonlinear patterns in the data. 
    
    In order to do this, we had to overcome several issues, most importantly, the overfitting towards identity.

    Traditionally,  overfitting towards identity is addressed by using dropout in the input layer \cite{Liang2018,Shenbin2020,Kim2019}. However, this approach is not effective enough and is not enabling the usage of really deep architectures.

    In this work, we propose three major contributions to address the issues mentioned above.  We propose:
    \begin{itemize}
    \item the usage of focal loss for training autoencoders for Top-N recommendation. 
    \item a simple yet effective data augmentation technique to prevent Top-N recommending autoencoders from overfitting towards identity. 
    \item a joint-learning technique based on the Hadamard product for training different combinations of various models. 
    \end{itemize}
    
    As a demonstration, we build the VASP, a Variational Autoencoder with a Shallow parallel Path. VASP combines deep Variational Autoencoder and a neural variant of shallow EASE jointly trained together to model both linear and non-linear patterns in the data. VASP was able to achieve state of the art performance on the MovieLens 20M dataset and competitive results on the Netflix Prize dataset.
  

\section{Related Work}


    
        Matrix Factorization (MF) has been the first choice model for many years since the team “BellKor’s Pragmatic Chaos” won the Netflix Prize\cite{Koren2009}. 
        
        In 2016 Sedhain et al. proposed AutoRec\cite{Sedhain2015}, an autoencoder-based model for collaborative filtering with explicit ratings that outperforms all current baseline models.
        
        After emergence of variational autoencoders (VAE), the collaborative filtering model MultVAE was proposed in 2018 by Liang et al.\cite{Liang2018} This model uses multinomial log-likelihood for data distribution. 
        
        H. Steck proposed EASE\cite{Steck2019}, the Embarrasingelly Shallow Autoencoder with no hidden layers as opposed to deep architectures. This approach was able to beat SOTA models when introduced.
        
        Several techniques for improving the MultVAE were proposed recently. RecVAE\cite{Shenbin2020} uses a separate regularization term in the form of the KL divergence between the actual parameter distribution and the distribution in previous training step preventing instability during training. 
        
        H+VAMP\cite{Kim2019} implements a variational autoencoder with a variational mixture of posteriors prior (Vamp Prior) with the goal to learn better latent representations of user-items interactions.
     

        During the evolution of recommender systems, many simple, shallow (linear, wide), and complex deep architectures have been proposed. Cheng et al. proposed a combination of those two approaches into a single framework called Wide\&Deep Learning \cite{Cheng2016} and introduced a technique called joint training. The authors also point out the distinction between joint training and ensembling. We are inspired by this work, but our approach is quite different. We do not process item attributes, just the interactions, therefore deep path is not design to encode items but to find nonlinear interaction patterns. Also, voting scheme is different.
        
        In\cite{drif2020ensvae}, the authors propose ensembling of pre-trained recommender models by variational autoencoder. However, joint-learning of such model seems to be problematic from the perspective of scaling and practical usability.
        
        Many other deep learning techniques originally developed for computer vision or natural language processing was later successfully used in other fields such as recommender systems. Residual networks \cite{He2016} are good example. Another example is using the approach from \cite{Huang2017} for dense layers in RecVAE \cite{Shenbin2020}.

        We follow this trend by adopting Focal Loss (FL) from \cite{Lin2020} for recommendation systems. This novel approach is used for imbalanced classes in object detection addressing the imbalance between the background class and other classes. That means that the loss is higher for examples in the training set that are difficult to classify. This perfectly fits the situation in collaborative filtering, where some items are more popular than others. It is more difficult to recommend niche items as they do not have many interactions. Higher loss for these items push recommender system to focus more on cold start and niche items. 

        Another essential idea while training an autoencoder on sparse data is to prevent overfitting towards learning identity function between the input and the output layer of the autoencoder\cite{Steck2020}. In \cite{zhang2017split}, the authors proposed the Split-Brain Autoencoder, which prevents learning identity by splitting input image into two separate channels: grayscale channel \(X_1\) and color channels \(X_2\). Learning is then performed in a separate way by training two networks, \(F_1\) to perform automatic colorization by learning \(X_2\) by showing \(X_1\) and \(F_2\) to make a grayscale prediction by learning \(X_2\) by showing \(X_1\).
        On the other hand, our approach uses only one neural network with automated data augmentation as a preprocessing step. 

\section{Our Approach}
        \begin{figure}
            \centering 
            \includegraphics[width=0.9\columnwidth]{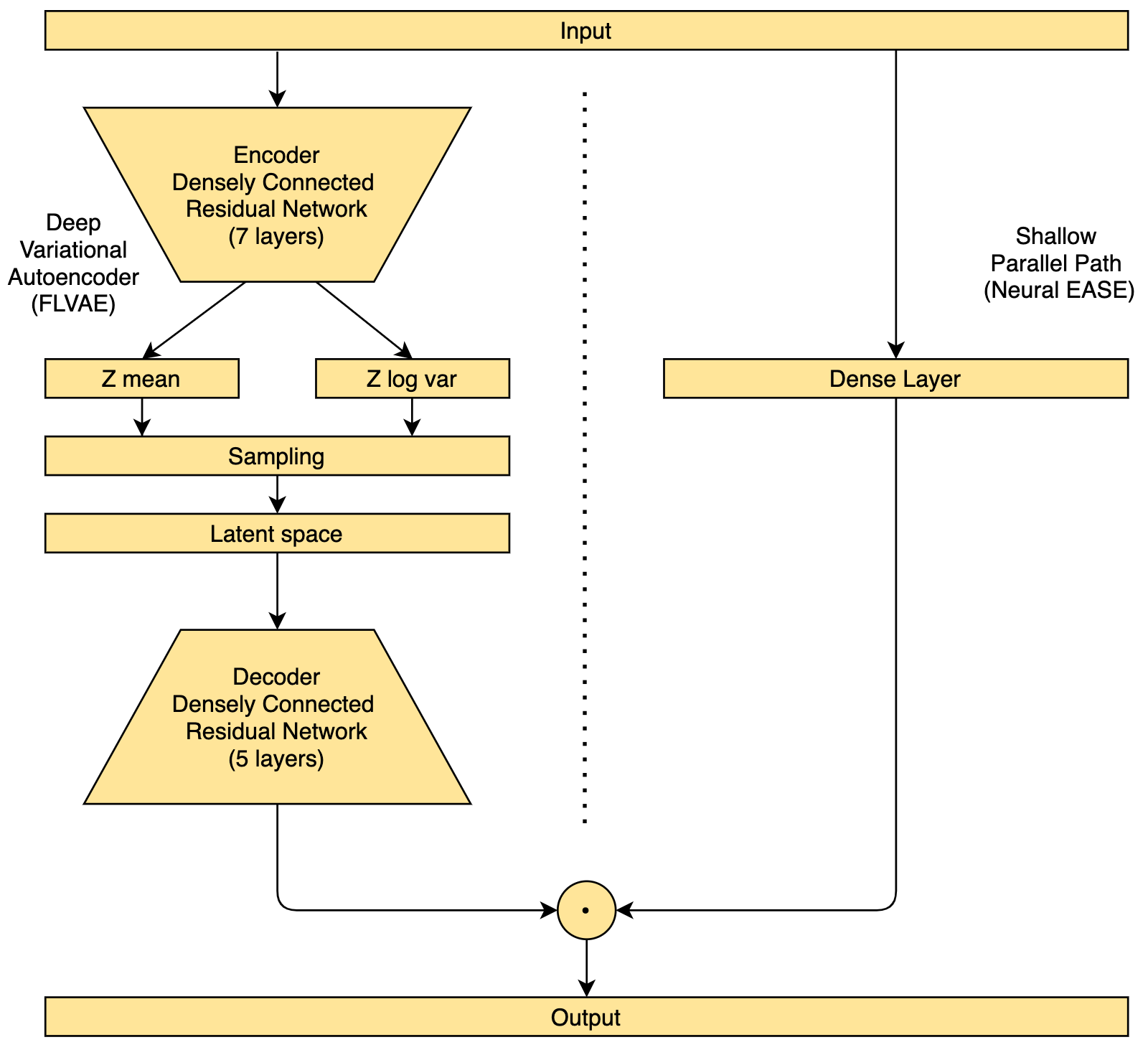} 
            \caption{VASP Architecture}
            \label{fig:vasp}
        \end{figure}  
    Following notation from\cite{Liang2018}, we index users as \(u\in\left\{1,...,U\right\}\), items as \(i\in\left\{1,...,I\right\}\), and user-item interaction matrix \(X \in \mathbb{N}^{U \times I}\). Lowercase \(\mathbf{x}_u=\left[ x_{u1}, x_{u2}, ... , x_{uI}  \right]^T \in \mathbb{N}^I\) denotes the interaction history and \(\mathbf{\hat{x}}_u=\left[ \hat{x}_{u1}, \hat{x}_{u2}, ... , \hat{x}_{uI}  \right]^T \in \mathbb{N}^I\) predicted ratings of the user \(u\).
    
    \subsection{Neural EASE (NEASE)}
        
        Following \cite{Steck2019}, the EASE model can be described as:
        
            \begin{equation}\label{ease_definition} 
                \mathbf{\hat{x}}_u= W \cdot \mathbf{x}_u,
            \end{equation}
        
        where \(W\in \mathbb{R}^{|I|\times{|I|}}\) is the weight matrix. Diagonal of \(W\) is constrained to zero to prevent learning identity function between the input and the output. In \cite{Steck2019} authors proposed using square loss between the data \(\mathbf{x}_u\) and the predicted scores \(\mathbf{\hat{x}_u}\) because this training objective has a closed-form solution. The authors also suggest that using more complex loss functions may lead to better prediction accuracy with higher computational costs.
        
        To enable running the model in parallel to a deep autoencoder, we interpret the EASE model (\ref{ease_definition}) as a single-layer perceptron  without bias nodes and with forced zeros on the diagonal, which can be trained with any suitable loss function using backpropagation.
        
        Our experiments with several different loss functions are described in Section \ref{results}.
         
    \subsection{MultVAE with focal loss (FLVAE)}

        Consistently with any other variational autoencoder \cite{Kingma2014}, the MultVae model's generative process starts by sampling \(k\)-dimensional latent representation \(\mathbf{z}_u\) from a standard Gaussian prior\cite{Liang2018}. 
        Then, under an assumption that interaction history \(\mathbf{x}_u\) has been drawn from a multinomial distribution, a neural network \(f_\theta(\cdot)\) is used to produce a probability distribution \(\pi(\mathbf{z}_u)\) over \(I\) items:
            \begin{gather*}\label{z_drawn}
                \mathbf{z}_u\sim N(0,\mathbf{I}_k), \\
                \pi(\mathbf{z}_u) \propto exp\{f_\theta(\mathbf{z}_u)\}, \\
                \mathbf{x}_u\sim Mult(N_u,\pi(\mathbf{z}_u))
            \end{gather*}

        Variational autoencoder then aims to maximize the average marginal likelihood  \(p(\mathbf{z}_u|\mathbf{x}_u)=\int_{}^{}p(\mathbf{x}_u|\mathbf{z}_u)p(\mathbf{z}_u)dz \). Since \(f_\theta(\cdot)\) is a neural network, \(p(\mathbf{z}_u|\mathbf{x}_u)\) becomes intractable and it is approximated with evidence lower bound (ELBO):
        
            \begin{equation}\label{elbo}
                \begin{aligned}
                    \log p \geq  
                    \mathbb{E}_{q}[\log p(\mathbf{x}_u|\mathbf{z}_u)-KL(q(\mathbf{z}_u|\mathbf{x}_u) || p(\mathbf{z}_u))]
                \end{aligned}
            \end{equation}
        
        where \(q(\mathbf{x}_u;\phi)\) is a variational approximation of the posterior
        distribution, \(p(\mathbf{x}_u;\theta)\) is the prior distribution, \(\phi\) and \(\theta\) are parameters of \(p(\mathbf{x}_u|\mathbf{z}_u)\), \(\log p(\mathbf{x}_u|\mathbf{z}_u)\) is the log-likelihood for user \(u\) and KL is the Kullback-Leibler divergence.
        
        FL\cite{Lin2020} is defined as 
        
            \begin{equation}\label{focal_loss} 
                FL(p_t)=-\alpha_t(1-p_t)^\gamma log(p_t),
            \end{equation} where \(p_t\) is:
        
            \begin{equation}\label{p_t} 
                p_t=\begin{cases}\hat{x}_{ui} & if \; x_{ui} = 1\\1 - \hat{x}_{ui}& otherwise\end{cases}
            \end{equation} and \(\alpha_t\), \(\gamma\) act as hyperparameters.
        
        Since maximising log-likelihood is the same as minimising cross-entropy and focal loss can be understood as a form of weighted cross-entropy, ELBO (\ref{elbo}) can be easily rewritten:
        
            \begin{equation}\label{elbo_fl}
                \begin{aligned}
                    \log p \geq &\mathbb{E}_{q}[\alpha_t(1-p(\mathbf{x}_u|z_u))^\gamma\log p(\mathbf{x}_u|\mathbf{z}_u) - \\
                               -&KL(q(\mathbf{z}_u|\mathbf{x}_u) || p(\textbf{z}_u))]
                \end{aligned}
            \end{equation}
    
    \subsection{VASP}\label{vasp}

        Recommender model \(m\) can be expressed as a function \(m(\cdot): \mathbf{x}_u \rightarrow \mathbf{\hat{x}}_u\). If \(m\) uses a sigmoid function on the output, it's obvious that \(\hat{x}_{uI} \in <0,1>\). 
        We propose joint-learning with Hadamard product\cite{million2007hadamard}, denoted \(\odot\) for combining any number \(n\), \(n \in \mathbb{N}\) of recommender models \(m_n\) as 
        
            \begin{equation}\label{hadamard}
                \begin{aligned}
                     m_n(\mathbf{x}_{u}) = \bigodot_{j=1}^n{m_j}=m_1(\mathbf{x}_{u}) \odot m_2(\mathbf{x}_{u})\odot...\odot m_n(\mathbf{x}_{u})
                \end{aligned}
            \end{equation}
        
        since \(m_n(\mathbf{x}_{u}) = \mathbf{\hat{x}}_{nu} \) and \(\mathbf{x}_{u}\) is the same for all models in combination, (\ref{hadamard}) can be directly rewritten as
        
            \begin{equation}\label{hadamard2}
                \begin{aligned}
                     m_n(\mathbf{x}_{u}) = \hat{x}_{nu} = \bigodot_{j=1}^n{\hat{x}_{ju}}
                \end{aligned}
            \end{equation}
        
        while
        
            \begin{equation}\label{hadamard_condition}
                \begin{aligned}
                    \hat{x}_{nu} \in \langle0,1\rangle.
                \end{aligned}
            \end{equation}
        
        While in Wide \& Deep\cite{Cheng2016}, networks are combined with the summation (logical OR), in VASP we use the Hadamard product (logical AND), meaning that both networks have to agree. In \cite{Cheng2016}, activation of a single network is sufficient for positive output.
        
        Proposed VASP architecture (Fig. \ref{fig:vasp}.) uses combination of two models, NEASE and FLVAE ensembled by element-wise multiplication (\ref{hadamard2}):
    
            \begin{equation}\label{hadamard3}
                \begin{aligned}
                     m_{VASP}(\mathbf{x}_{u}) &= m_{FLVAE}(\mathbf{x}_{u}) \odot m_{EASE}(\mathbf{x}_{u}) \\
                                     &= \mathbf{\hat{x}}_{FLVAEu} \odot \mathbf{\hat{x}}_{NEASEu}
                \end{aligned}
            \end{equation}
        
        To satisfy condition (\ref{hadamard_condition}) we add sigmoid function to (\ref{ease_definition}):
            
            \begin{equation}\label{nease_definition} 
                \mathbf{\hat{x}}_{NEASEu} = \sigma(W \cdot \mathbf{x}_u)
            \end{equation}
        
        Since \(m_{FLVAE}\) and \(m_{EASE}\) are both fully differentiable, \(m_{VASP}\) is also fully differentiable and backpropagation can be used to optimize the \(m_{VASP}\).
        
    \subsection{Data augmentation to prevent learning identity}\label{pridentity} 
        \begin{figure*}[hb]
            \centering 
            \includegraphics[width=1.9\columnwidth]{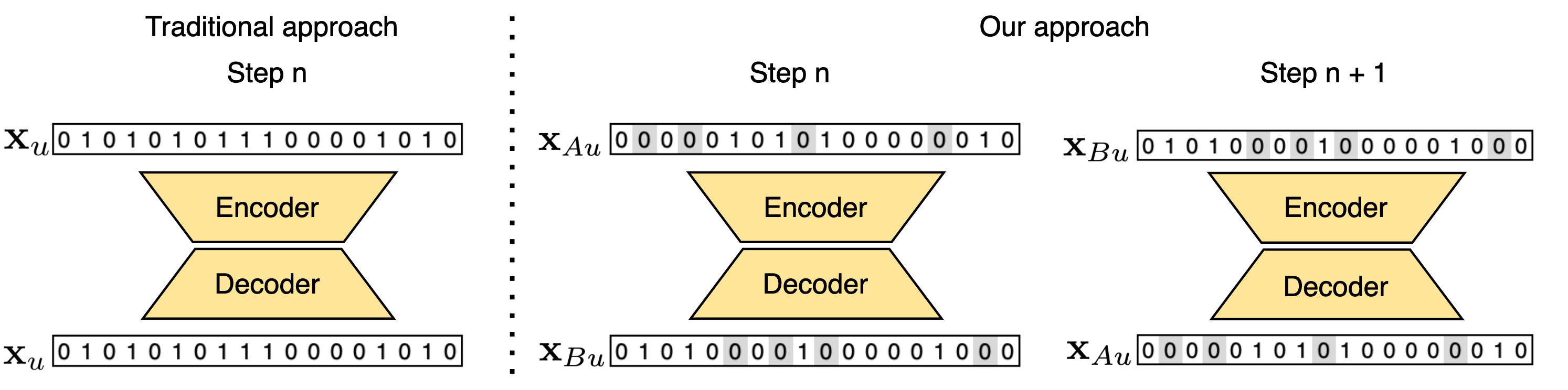} 
            \caption{Data augmentation to prevent overfitting towards identity}
            \label{fig:vasp_identity}
        \end{figure*}
        
        Inspired by \cite{zhang2017split} we prevent learning identity by spliting the input interactions \(x_u\) before every training epoch randomly into two parts, \(\mathbf{x}_{Au}\) and \(\mathbf{x}_{Bu}\) so:
        
            \begin{equation}\label{p_t} 
                \begin{aligned}
                    x_{Aui}=\begin{cases}0 & if \; x_{ui} = 0\\1-x_{Bui}& otherwise\end{cases} \\
                    x_{Bui}=\begin{cases}0 & if \; x_{ui} = 0\\1-x_{Aui}& otherwise\end{cases}
                \end{aligned}
            \end{equation} while 
            \begin{equation}
                    \sum_{i=1}^I{x_{Aui}} \approx \sum_{i=1}^I{x_{Bui}}
            \end{equation}

        The autoencoder is learning \(x_{Bui}\) by showing \(x_{Aui}\) in one training step and than \(x_{Aui}\) by showing \(x_{Bui}\) in another.  Thus autoencoder is still seeing all the data, but does not see the identity and cannot learn it (as demonstrated by Figure \ref{fig:vasp_identity}).
            
    
\section{Interpreting VASP}
In order to analyze the effect of shallow and deep components of the proposed ensemble model, we have implemented a workflow to produce a visualization of movie embeddings learned by individual models and the ensemble.

\begin{figure*}
            \centering 
            \includegraphics[width=\linewidth]{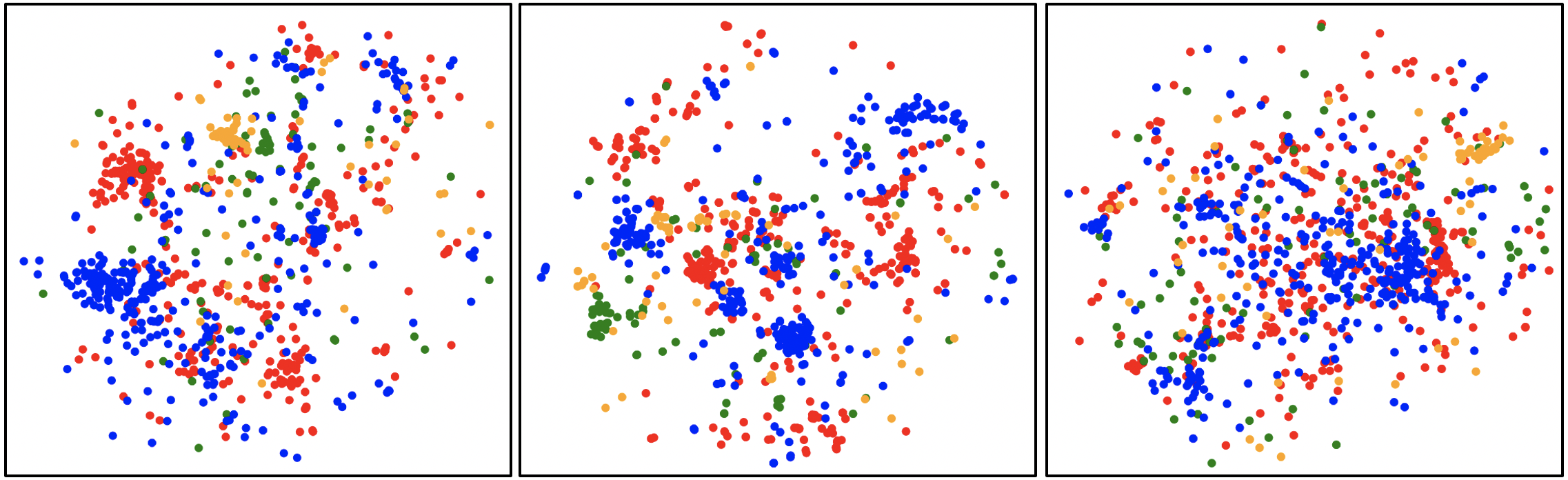} 
            \caption{Explaining VASP on MovieLens20M Dataset: Output of the joint model (left) was linearly decomposed to EASE component (middle) and FLVAE component (right) to demonstrate that EASE is learning more apparent linear dependencies and FLVAE the non-linear ones. Red = Horrors, Blue = Children movies, Green = Western movies and Yellow = Noir.
            }
            \label{fig:explain_vasp}
    \end{figure*}

The principle was the following. We performed a sensitivity analysis of models by putting one-hot vectors to the input and generating output probabilities (reconstructions). These probabilities were then transformed to distances by linear scaling and then projected into a t-SNE plot. See Figure \ref{fig:explain_vasp} for further details.
   
\section{Experimental Setup}\label{expsetup}
  
    To verify our assumtions by experiments, we implemented three models: neural variant of EASE, deep Variational Autoencoder, and then VASP: joint learning model consisting of NEASE and deep Variational Autoencoder ensembled by the Hadamard product as described in chapter \ref{vasp}. We trained those models on two datasets: MovieLens20M and Netflix prize dataset, and compared results over various baselines, including current SOTA models.
    
    \subsection{Datasets} 
        
        
        \subsubsection{MovieLens20M \cite{movieLens}} Dataset of 27000 movies rated by 138,000 users generating 20 million ratings in total. We preprocessed the dataset according to \cite{Liang2018}: Since this dataset contains explicit ratings, we converted the data to implicit interactions by considering valid interaction only rating of four or higher\footnote{Note that we use implicit interactions because it is prevalent case for practical recommendation tasks}. Only users with five or more interactions remain in the dataset after preprocessing. We randomly choose 10000 users as a test set and train our models on the rest.
        
        \subsubsection{Netflix Prize Dataset \cite{Bennett2007}} Dataset from Netflix prize - over 100 million ratings from 480000 randomly-chosen, anonymous Netflix customers over 17000 movie titles. We converted explicit ratings to implicit interactions by the same method as we used on the MovieLens20M. We randomly choose 40000 users as a test set and train our models on the rest.
        
        
    \subsection{Metrics}
        
        We evaluate our models in the same way as in \cite{Liang2018}. First, we sample 80\% of the test user's interactions as input for the model, and then we measure \(Recall@k\) and \(NCDG@k\) for predicted interactions against the remaining 20\% of the user's interactions. 
        
        \(NCDG\) for TOP-k recommended items, denoted \(NCDG@k\) is defined as 
        
        \begin{equation}\label{ndcg} 
            NDCG@k = \frac{DCG@k}{IDCG@k}
        \end{equation}
        
        where 
        
        \begin{equation}\label{dcg} 
            DCG@k = \sum_{i=1}^{k} \frac{2^{rel_{i}}-1}{log_2(i+1)}
        \end{equation}
        
        and     
        
        \begin{equation}\label{idcg} 
            IDCG@k = \sum_{i=1}^{|R_k|} \frac{2^{rel_{i}}-1}{log_2(i+1)}
        \end{equation}
        
        \(rel_{i}\) is relevance of the recommendation at position \(i\) and \(R_k\) is the list of those 20\% interactions acting as "true" user interactions.
        
        \(Recall\) for TOP-k recommended items, denoted \(Recall@k\) is defined as
        
        \begin{equation}\label{recall} 
            Recall@k = \frac{|\hat{R_k} \cap R_k|}{R_k}
        \end{equation}
        
        where \(\hat{R_k}\) is top-k items recommended by evaluated model.
        
    \subsection{Baselines}
        
        We chose several autoencoder-based models for a top-n recommendation, including MultDAE and MultVAE from \cite{Liang2018}, EASE from \cite{Steck2019} and current SOTA models, RecVAE \cite{Shenbin2020} and H+VAMP \cite{Kim2019} as baselines for performance evaluation.

    \subsection{Implemented models}
        
        \subsubsection{Neural EASE} was implemented as a dense layer with forced zeros on diagonal by kernel constraint. We evaluate three different loss functions: mean squared error, cosine proximity loss and focal loss. Since EASE has forced zeros on diagonal in parametrs matrix to prevent learning identity, no data augmentation was used.
    
        \subsubsection{Variational Autoencoder} we implemented FLVAE  with densely connected residual network both in encoder and decoder. Sigmoid activation was used on the output. 
        
        
        Data augmentation described in section \ref{pridentity} was used to prevent autoencoder in learning identity.  
        
        Default values of \(\alpha_t=0.25\) and  \(\gamma=2.0\) in (\ref{focal_loss}) was used.
        
        \subsubsection{VASP} is build by connecting neural EASE and FLVAE by the Hadamard product as described in \ref{vasp}. Hyperparameters was the same as for plain FLVAE.
        We evaluated three variants: pre-trained EASE and FLVAE joined together as ensemble, jointly training form the start and alternating approach where FLVAE and EASE was training in every step separatedly. We prevent our model to learn identity by using data augmentation described in \ref{pridentity}.
        
        \subsection{Hyperparameters}
        We used 2048 units for latent space and 4096 units for hidden layers. We used seven densely connected residual hidden layers for the encoder and five layers of the same architecture in the decoder.
        
        We trained our model for 50 epochs with a learning rate of 0.00005 and batches of 1024 samples. Then, we lower the learning rate to 0.00001 and trained for another 20 epochs. Then we performed finetuning with a learning rate of 0.000001 for another 20 epochs.
        
    All models was implemented in Tensorflow\cite{tensorflow2015-whitepaper} and the source code with notes for reproducing the results is publicly available on our \href{https://github.com/zombak79/vasp}{GitHub page}.
    
\section{Results and Discussion}\label{results}

    \textbf{Neural EASE}: We evaluated three different variants of the model based on the loss function used - mean squared error, cosine loss, and focal loss (see Table \ref{tab:2}).
    
    \begin{table}[h]\label{tab:2}\caption{Results with different loss functions for the EASE model on MovieLens20M dataset.}
        \centering
        \begin{tabular}{@{}llll@{}}
            \toprule
            Loss function used & NCDG@100 & Recall@20 & Recall@50 \\ 
            \midrule
            MSE                & \multicolumn{1}{c}{0.425} & \multicolumn{1}{c}{0.393} & \multicolumn{1}{c}{0.523}           \\
            Cosine proximity & \multicolumn{1}{c}{0.431} & \multicolumn{1}{c}{0.403} & \multicolumn{1}{c}{0.532} \\
            Focal loss         &   \multicolumn{1}{c}{0.377} & \multicolumn{1}{c}{0.343} & \multicolumn{1}{c}{0.426}             \\ 
            \bottomrule
        \end{tabular}
        \centering
    \end{table}
    
    \noindent
    We found that using cosine proximity loss leads to better performance of the model as authors of \cite{Steck2019} expected.
    
    \textbf{FLVAE}: Authors of MultVAE reject deeper architectures by stating that "going deeper does not improve performance." We investigated the matter and believed that overfitting towards identity was to blame. We address this issue by adopting data augmentation described in the section \ref{pridentity}. This approach successfully allowed us to build much bigger models than in \cite{Liang2018}, \cite{Shenbin2020} or \cite{Kim2019} where 200 units for latent space and 600 in hidden layers was used.
    
 
    \textbf{VASP}: We evaluated three methods of training models connected by the Hadamard product (see Table \ref{tab:3}). 
    
    \begin{table}[h]\label{tab:3}\caption{Results with different training approach for the VASP model on MovieLens20M dataset.}
        \centering
        \begin{tabular}{@{}llll@{}}
            \toprule
            Training approach & NCDG@100 & Recall@20 & Recall@50 \\ 
            \midrule
            Pretrained ensemble  & \multicolumn{1}{c}{0.442} & \multicolumn{1}{c}{0.414} & \multicolumn{1}{c}{0.545}           \\
            Alternating training & \multicolumn{1}{c}{0.436} & \multicolumn{1}{c}{0.401} & \multicolumn{1}{c}{0.543} \\
            Joint learning &   \multicolumn{1}{c}{0.448} & \multicolumn{1}{c}{0.414} & \multicolumn{1}{c}{0.552}             \\ 
            \bottomrule
        \end{tabular}
        \centering
    \end{table}

    First, we connected pre-trained models and evaluated them as an ensemble. Then we initialized the joint model and trained it from scratch. Lastly, we experimented with the alternating approach, where in one step was frozen weights of FLVAE, and in the next step, we weights of the EASE model were frozen instead. However, this approach did not perform better than joint-learning from the start.

%
%
%
        
        
    
    \begin{table*}\label{tab:1}\caption{Results}
        \centering
        \begin{tabular}{@{}lcccccc@{}}
            \toprule
                            & \multicolumn{3}{c}{MovieLens 20M}& \multicolumn{3}{c}{Netflix Prize Dataset} \\ 
            \midrule
                    \multicolumn{1}{l}{} & NCDG@100 &Recall@20 & Recall@50 & NCDG@100 & Recall@20 & Recall@50 \\ 
            \midrule 
                \multicolumn{1}{l}{Mult-DAE}     & 0.419 & 0.387 & 0.524 & 0.380 & 0.344 & 0.438 \\
                \multicolumn{1}{l}{Mult-VAE}     & 0.426 & 0.395 & 0.537 & 0.386 & 0.351 & 0.444 \\
                \multicolumn{1}{l}{EASE}         & 0.420 & 0.391 & 0.521 & 0.393 & 0.362 & 0.445 \\
                \multicolumn{1}{l}{RecVAE}       & 0.442 & {\bf 0.414} & {\bf 0.553} & 0.394 & 0.361 & 0.452 \\
                    \multicolumn{1}{l}{H+Vamp Gated} & 0.445 & {\bf 0.413} & 0.551 & {\bf 0.409} & {\bf 0.376} & {\bf 0.463} \\ 
            \midrule 
                \multicolumn{1}{l}{Neural EASE}        & 0.431 & 0.403 & 0.532 & 0.395 & 0.363 & 0.447 \\
                \multicolumn{1}{l}{FLVAE}        & 0.445 & 0.409 & 0.547 & 0.398 & 0.363 & 0.450 \\
                \multicolumn{1}{l}{VASP}         & {\bf 0.448} & {\bf 0.414} & {\bf 0.552} & 0.406 & 0.372 & 0.457 \\ 
            \bottomrule
        \end{tabular}

    \end{table*}
  
  Finally, we have compared our base models NEASE, FLVAE and jointy learned ensemble VASP to state of the art approaches (see Table \ref{tab:1}). Significantly best performing models are in bold. Our VASP outperformed other models and achieved the SOTA for MovieLens 20M dataset. It also performed quite well for the Netflix dataset (second highest ranking model). 
  
  In our future experiments, we will carefully analyse the H+Vamp Gated model that performed better on Netflix. We will try to put it in the ensemble with the Neural EASE or use the idea of Variational Mixture of
Posteriors to improve performance of our deep FLVAE model.

\section{Conclusion}
    
    We proved EASE to be a compelling Top-N recommendation model that can still match current SOTA baselines.
    
    We proposed a data augmentation method to prevent overfitting to identity and experimentally proved that using this method leads to better performance of autoencoders used for top-n recommendation.
    
    We proposed a novel joint-learning technique for training multiple models together. Using that we constructed VASP, Variational Autoencoder with parallel Shalow Path and experimentally proved, that variational autoencoder connected with parallel simple shallow linear model can match current sophisticated SOTA models and even outperform them in some cases.

\section*{Acknowledgment}
Our research has been supported by the Grant Agency of the Czech Technical University in Prague (SGS20/213/OHK3/3T/18), the Czech Science Foundation (GA\v{C}R 18-18080S), Recombee and VUSTE-APIS.


\bibliographystyle{plain} 
\bibliography{bibliography} 

\end{document}